\documentclass[letterpaper]{article} 
\usepackage{aaai24}  
\usepackage{times}  
\usepackage{helvet}  
\usepackage{courier}  
\usepackage[hyphens]{url}  
\usepackage{graphicx} 
\usepackage{natbib}  
\usepackage{caption} 
\usepackage{algorithm}
\usepackage{algorithmic}
\usepackage{multirow}
\usepackage{booktabs}

\usepackage{amsmath,amsfonts,bm}

\def\eqref#1{equation~\ref{#1}}

\def\1{\bm{1}}

\def\vmu{{\bm{\mu}}}

\def\vb{{\bm{b}}}

\def\vx{{\bm{x}}}
\def\vy{{\bm{y}}}

\def\veps{{\bm{\epsilon}}}

\def\vdelta{{\bm{\delta}}}

\def\mI{{\bm{I}}}

\DeclareMathAlphabet{\mathsfit}{\encodingdefault}{\sfdefault}{m}{sl}
\SetMathAlphabet{\mathsfit}{bold}{\encodingdefault}{\sfdefault}{bx}{n}

\def\sR{{\mathbb{R}}}

\def\ttp{{\tt p}}
\def\ttP{{\tt P}}

\def\sE{{\mathbb{E}}}

\usepackage{amsthm}
\newtheorem{theorem}{Theorem}
\newtheorem{thm}{Theorem}
\newtheorem{proposition}{Proposition}
\newtheorem{lemma}{Lemma}
\newtheorem{definition}{Definition}
\newtheorem{assumption}{Assumption}
\newtheorem{parametersetting}{Parameter Setting}
\usepackage{xcolor}
\usepackage{newfloat}
\usepackage{listings}
\DeclareCaptionStyle{ruled}{labelfont=normalfont,labelsep=colon,strut=off} 
\floatstyle{ruled}
\newfloat{listing}{tb}{lst}{}
\floatname{listing}{Listing}
\title{SimCalib: Graph Neural Network Calibration based on Similarity between Nodes}
\author{
    Boshi Tang\textsuperscript{\rm 1}, Zhiyong Wu\textsuperscript{\rm 1}, Xixin Wu\textsuperscript{\rm 2}\thanks{Corresponding author.},\\ Qiaochu Huang\textsuperscript{\rm 1}, Jun Chen\textsuperscript{\rm 1}, Shun Lei\textsuperscript{\rm 1}, Helen Meng\textsuperscript{\rm 2}
}
\affiliations{
    \textsuperscript{\rm 1}Shenzhen International Graduate School, Tsinghua University, Shenzhen, China\\
    \textsuperscript{\rm 2}The Chinese University of Hong Kong, Hong Kong SAR, China\\
    tbs22@mails.tsinghua.edu.cn, zywu@sz.tsinghua.edu.cn, wuxx@se.cuhk.edu.hk,\\
    \{hqc22, y-chen21, leis21\}@mails.tsinghua.edu.cn, hmmeng@se.cuhk.edu.hk
}

\begin{document}

\maketitle

\begin{abstract}
Graph neural networks (GNNs) have exhibited impressive performance in modeling graph data as exemplified in various applications. Recently, the GNN calibration problem has attracted increasing attention, especially in cost-sensitive scenarios. Previous work has gained empirical insights on the issue, and devised effective approaches for it, but theoretical supports still fall short. In this work, we shed light on the relationship between GNN calibration and nodewise similarity via theoretical analysis. A novel calibration framework, named \textit{SimCalib}, is accordingly proposed to consider similarity between nodes at global and local levels. At the global level, the Mahalanobis distance between the current node and class prototypes is integrated to implicitly consider similarity between the current node and all nodes in the same class. At the local level, the similarity of node representation movement dynamics, quantified by nodewise homophily and relative degree, is considered. Informed about the application of nodewise movement patterns in analyzing nodewise behavior on the over-smoothing problem, we empirically present a possible relationship between over-smoothing and GNN calibration problem. Experimentally, we discover a correlation between nodewise similarity and model calibration improvement, in alignment with our theoretical results. Additionally, we conduct extensive experiments investigating different design factors and demonstrate the effectiveness of our proposed SimCalib framework for GNN calibration by achieving state-of-the-art performance on 14 out of 16 benchmarks.

\end{abstract}

\section{Introduction}
Graphs are ubiquitous in the real world, including social networks, knowledge graphs, traffic networks, among others. Due to the universality and expressive power of graph representations, the deep learning community has paid much attention to learning from graph-structured data and introduced various types of graph neural networks (GNNs) \cite{kipf2016semi, velivckovic2017graph, hamilton2017inductive}. To date, GNNs have been successfully applied to various downstream applications with remarkable accuracy, such as drug discovery \cite{zhang2022graph}, fluid simulation \cite{liu2022fluid}, and recommendation system \cite{fan2019graph}, to name a few.

However, in many applications, trustworthiness is as important (if no more) than accuracy, especially in safety-sensitive fields \cite{dezvarei2023graph}. One of the promising solution to ensure trustworthiness of a trained model is aligning its prediction confidence with the ground truth accuracy, i.e., the model should provide the appropriate confidence to reveal whether the prediction should be trusted. Unfortunately, such an alignment is hardly achieved by modern neural networks \cite{guo2017calibration,wang2021confident}. To mitigate the issue, a variety of calibration methods \cite{kull2019beyond, gupta2020calibration, zhang2020mix} have been proposed to calibrate pretrained deep neural networks (DNNs). However, calibration for GNNs is still underexplored. While it is possible to directly apply calibration methods designed for DNNs to GNNs by treating each node in the graph as an isolated sample, the specific challenges posed by GNN calibration remain unaddressed as the specific characteristics of graph structure, e.g., the relationship between nodes, is not well utilized for calibrating the GNN predictions.

Recently, a few works focus on GNN calibration, among which the most noticeable are CaGCN \cite{wang2021confident} and GATS \cite{teixeira2019graph}. Specifically, CaGCN produces nodewise temperatures by processing the pretrained classifier's logits with another graph convolutional network (GCN), in the hope that structural information can be implicitly integrated in model calibration. Following it, GATS empirically investigates factors that influence GNN calibration, and employs an attention network to account for the influential factors. However, to date, the efforts towards such a structured prediction problem \cite{nowozin2011structured} have mostly concentrated on empirical aspects, suffering from a lack of theoretical supports.

In this paper, we make extensive efforts from both theoretical and practical aspects to tackle the aforementioned issues. Our main contributions are summarized as follows:
\begin{itemize}
    \item We develop a theoretical approach for GNN calibration, and prove that by taking nodewise similarity into consideration we can reduce expected calibration error (ECE) effectively.
    \item We propose two similarity-oriented mechanisms to account for both global feature-level similarity and local nodewise representation movement dynamics similarity. By incorporating them into network designs, we propose SimCalib, a novel GNN calibration method that is data-efficient, easy to implement and highly expressive.
    \item We are the first to relate the oversmoothing problem to GNN calibration.
    \item We conduct comprehensive experiments investigating various design factors, and demonstrate the effectiveness of SimCalib by achieving new SOTA performance on 14 out of 16 benchmarks. Particularly, compared with the previous SOTA model, SimCalib on average reduces ECE by 10.4\%.
\end{itemize}

\section{Preliminary}
\subsection{Problem setting}
Herein we consider the problem of calibrating GNNs on semi-supervised node classification tasks. Specifically, given an undirected graph $\mathcal{G}=(\mathcal{V}, \mathcal{E})$, consisting of nodes $\mathcal{V}$ and edges $\mathcal{E} \subseteq \mathcal{V} \times \mathcal{V}$, each node $v_i \in \mathcal{V}$ is associated with a feature vector $\vx_i \in \mathbb{R}^d$. Moreover, a proper subset of nodes, denoted as $\mathcal{L} \subset \mathcal{V}$, is further associated with labels $\{y_i\}_{i : v_i \in \mathcal{L}}$, where $y_i \in \mathcal{Y} = \{1,\dots,K\}$ is the ground-truth label for $v_i$. And the goal of semi-supervised node classification is to infer the labels for the unlabeled nodes $\mathcal{U} = \mathcal{V} \setminus \mathcal{L}$. A graph neural network approaches the problem by taking into account both nodewise features and structural information, i.e. adjacency matrix $\mathbf{A}$, and it predicts a probability distribution $\hat{p}_i$ over all the classes for each node $v_i$. The value on the $j$-th position of the distribution, i.e. $\hat{p}_i^{(j)}$, describes the estimated probability of $v_i$ being in class $j$.

For each node, $\hat{p}_i$ induces the corresponding label prediction $\hat{y}_i := {\tt \mathop{argmax}}_j \ \hat{p}_i^{(j)}$ and confidence $\hat{c}_i := {\tt \mathop{max}}_j \ \hat{p}_i^{(j)}$. Perfect calibration is defined as \cite{wang2021confident}:
\begin{equation}
    \forall \ c\in[0, 1],\quad {\tt P}(y_i=\hat{y}_i|\hat{c}_i=c)=c. \label{eq:perf_cal}
\end{equation}
In practice, perfect calibration cannot be estimated with a finite number of samples, therefore calibration quality is often quantified by expected calibration error (ECE) instead \cite{naeini2015obtaining, guo2017calibration}:
\begin{eqnarray}
    {\tt ECE}:=\sE_{p}\Bigg[\Bigg|\sE[Y=k|\tt\hat{p}(Y=k|\vx)=p]-p\Bigg|\Bigg],
\end{eqnarray}
where ${\tt \hat{p}}(Y=k|\vx)$ is the predicted probability of $\vx$ being in class $k$, the inner expectation represents the ground-truth probability of $\vx$ belonging to $k$, and the outer expectation iterates over all $p\in (0,1)$.

\subsection{Nodewise temperature}
To preserve the nodewise predictions, CaGCN calibrates logits by scaling them with nodewise temperatures, i.e. 
\begin{equation}
    \hat{z}_i' = \frac{\hat{z}_i}{T_i},
\end{equation}
where $\hat{z}_i$ is the nodewise logits for $v_i$ produced by the pretrained classifier and $T_i > 0$ is the temperature for $v_i$ estimated by CaGCN. A noticeable property of such a mechanism is
\begin{equation}
    \mathop{{\tt argmax}}_{j\in \mathcal{Y}}\ \hat{z}_i' = \mathop{{\tt argmax}}_{j\in \mathcal{Y}}\ \hat{z}_i,
\end{equation}
which maintains the prediction accuracy of GNNs after calibration. In this work, we follow the practice and calibrate models in the same manner.

\section{Theoretical Results}
We consider the Gaussian mixture block model\cite{li2023spectral}, which is commonly used for theoretical analysis on graphs and neural network calibration\cite{carmon2019unlabeled,zhang2022and}. 
\begin{definition} (Gaussian model). For $\vmu\in\sR^d$ and $\sigma>0$, the Gaussian model is defined as a distribution over $(\vx,y)\in\sR^d\times\{-1,1\}$:
\begin{eqnarray}
    \vx|y\sim\mathcal{N}(\vmu \cdot y,\sigma^2\mI),
\end{eqnarray}
where $y$ follows the Bernoulli distribution $\ttP(y=1)=\ttP(y=-1)=1/2$.
\end{definition}
\begin{assumption}
For two graph nodes $i$ and $j$ with the Gaussian model parameterized by $\vx_i|y_i\sim\mathcal{N}(\vmu_i\cdot y_i,\sigma^2\mI), \vx_j|y_j\sim\mathcal{N}(\vmu_j\cdot y_j,\sigma^2\mI)$, there exists a underlying linear relationship between the two nodes: $\vmu_j=a\vmu_i+\vb$, where $a\in \sR, \vb\in\sR^d$ are constants, and $||\vmu||^2=d$.
\end{assumption}

\begin{parametersetting}
    We choose the model parameters that allow a classifier with non-trivial standard accuracy (e.g., $\geq 1\%$) to be learned with high probability, following the Theorem 4 of \cite{schmidt2018adversarially}:
    \begin{eqnarray}
        ||\vmu||^2=d,\frac{||\vmu||^2}{\sigma^2}=\sqrt{\frac{d}{n}}\gg \frac{1}{\epsilon^2},\epsilon\in(0,\frac{1}{2})
    \end{eqnarray}
\end{parametersetting}

Given $n$ graphs as training samples $\{\vx^{(k)},y^{(k)}\}_{k=1...n}$, the estimator for Gaussian distribution parameters of node $i$ based on likelihood can be obtained as:
\begin{eqnarray}
    \bar{\vmu}_i=\frac{1}{n}\sum_{k=1}^n\vx_i^{(k)}y_i^{(k)}.
\end{eqnarray}
If nodes $i$ and $j$ are considered jointly, the estimator can then be obtained as:
\begin{eqnarray}
    \hat{\vmu}_i &=& \frac{1}{n(1+a^2)}\sum_{k=1}^n[y_i^{(k)}\vx_i^{(k)}+a(y_j^{(k)}\vx_j^{(k)}-\vb)] \\
    &=& \frac{1}{1+a^2}\bar{\vmu}_i+\frac{a^2}{1+a^2}\frac{\bar{\vmu}_j}{a}-\frac{a\vb}{1+a^2}. \label{eq:original_estimator}
\end{eqnarray}

Then, considering the ECE measure for the two estimators $\hat{\vmu}$ and $\bar{\vmu}$, 
\begin{eqnarray}
    {\tt ECE}_{\hat\vmu}&=&\sE_{p}\Bigg[\Bigg|\sE[Y=1|\tt\hat{p}(Y=1|\vx)=p]-p\Bigg|\Bigg]\\
    &=&\sE_{v=\hat{\vmu}_i^\top\vx}\Bigg[\Bigg|\frac{1}{e^{-\frac{2\hat{\vmu}^\top\vmu}{||\hat{\vmu}||^2}v}+1}-\frac{1}{e^{-2v}+1}\Bigg|\Bigg], \label{eq:joint}\\
    {\tt ECE}_{\bar\vmu}&=&\sE_{v=\bar{\vmu}_i^\top\vx}\Bigg[\Bigg|\frac{1}{e^{-\frac{2\bar{\vmu}^\top\vmu}{||\bar{\vmu}||^2}v}+1}-\frac{1}{e^{-2v}+1}\Bigg|\Bigg], \label{eq:isolated}
\end{eqnarray}
we have the following theorem for the expected cost minimizing (ECM) classifier defined in App. Prop. 1:
\begin{theorem}
    Under the above parameter setting, there exist numerical constants $c_0,c_2$, with $d/n>c_0$ and $a^2>(\frac{d}{n})^{1/4}/2$,
\begin{eqnarray}
    {\tt ECE}_{\hat\vmu} \leq {\tt ECE}_{\bar\vmu} \text{with probability}~\geq 1-e^{-c_2d/32}.
\end{eqnarray}
\end{theorem}
\begin{proof} We defer the detailed proof to the appendix. Here we give a sketch of the proof.
According to Lemma 2-5 (as proved in the appendix), with sufficiently large $d/n>c_0$ and high correlation $a^2$,
\[
    \ttp\Bigg(1\geq\frac{\hat{\vmu}^\top\vmu}{||\hat{\vmu}||^2}\geq \frac{\bar{\vmu}^\top\vmu}{||\bar{\vmu}||^2}\geq\frac{1}{2}\Bigg) \geq 1-e^{-c_2d/32},
    \]
    and $\frac{1}{e^{-\frac{2\hat{\vmu}^\top\vmu}{||\hat{\vmu}||^2}v}+1}$ is always closer to $\frac{1}{e^{-2v}+1}$ than $\frac{1}{e^{-\frac{2\bar{\vmu}^\top\vmu}{||\bar{\vmu}||^2}v}+1}$ with various $v$. Thus, 
    \[
    \ttp({\tt ECE}_{\hat\vmu} \leq {\tt ECE}_{\bar\vmu})\geq 1-e^{-c_2d/32}.
    \]
\end{proof}
The above theorem indicates that by jointly considering nodes with high correlation, the calibration error can be reduced effectively. This motivates our design of SimCalib which explicitly considers the similarity between nodes at both global and local levels.

\section{Related Work}
\subsection{Calibration for standard multi-class classification}
The model calibration task was first proposed in 2017 \cite{guo2017calibration}. About this problem, works can be roughly classified as post-hoc and training based methods. Post-hoc methods calibrate pretrained nodewise classifiers in ways that preserve predictions, as featured by temperature scaling (TS) \cite{guo2017calibration}, ensemble temperature scaling (ETS) \cite{zhang2020mix}, multi-class isotonic regression (IRM) \cite{zhang2020mix}, spline calibration \cite{gupta2020calibration}, Dirichlet calibration \cite{kull2019beyond}, etc. In contrast, instead of transforming logits from a pretrained classifier, training based methods modify either the model architecture or the training process itself. A plethora of methods based on evidential theory \cite{sensoy2018evidential}, model ensembling \cite{lakshminarayanan2017simple}, adversarial calibration \cite{tomani2021towards} and Bayesian approach \cite{hernandez2015probabilistic, wen2018flipout} belongs to the category. Whereas training based methods provide more flexibility compared to post-hoc ones, a limitation is that they hardly promise accuracy-preserving, thereby requiring careful trade-off between accuracy and calibration performance.
\subsection{GNN calibration}
Comparatively, GNN calibration is currently less explored. Teixeira et al.\shortcite{areGNNMis} empirically evaluate the post-hoc model calibration techniques developed for the standard i.i.d. setting on GNN calibration, and show that such a paradigm fails in the task due to an oversight of graph structural information. Afterwards, CaGCN \cite{wang2021confident} produces nodewise temperatures by processing nodewise logits via a graph convolutional network, to account for the graph structure. Additionally, GATS \cite{teixeira2019graph} experimentally points out influential factors of GNN calibration and produces nodewise temperatures with an attention-based architecture. Furthermore, Hsu et al.\shortcite{hsu2022graph} propose edgewise calibration metrics. Recently, uncertainty quantification is also considered via conformal prediction \cite{huang2023uncertainty, zargarbashi2023conformal}. However, our post-hoc calibration strategy differs from all the previous works with theoretical foundation and similarity-oriented mechanisms.

\begin{figure}[t]
    \includegraphics[width=9cm]{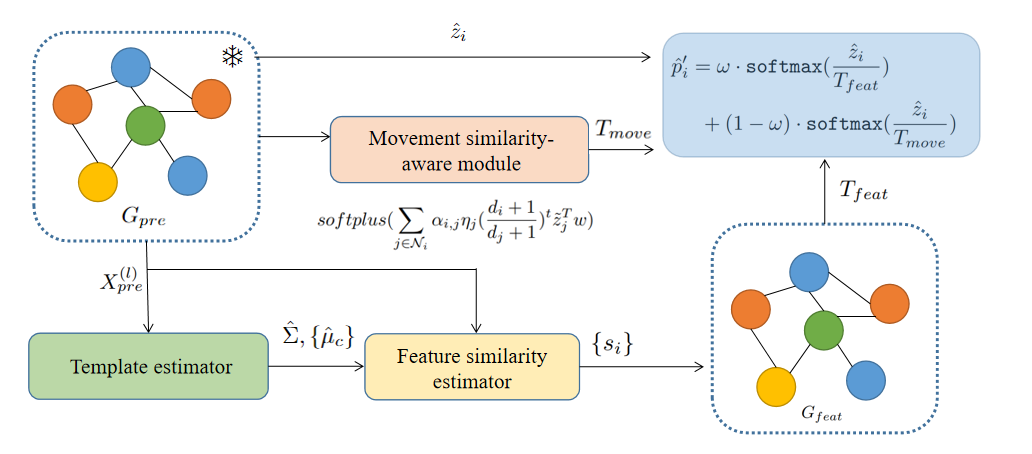}
    \caption{Figure of our proposed calibration pipeline. Note that the pretrained classifier is frozen during the calibration process.}
    \label{fig:model}
\end{figure}

\section{Methods}
In light of our theorem, we aim at exploiting nodewise similarity in the process of GNN calibration. Thus we discuss two mechanisms, namely feature and representation movement similarities in this section.

\subsection{Feature similarity propogation}
An intuitive form of feature similarity is raw feature similarity. However, we find with experiments that it aligns badly with GNN predictions, and also the subsequent calibration process, thereby performing suboptimally in GNN calibration. Thus we calculate similarity based on intermediate features from the pretrained GNN classifier $G_{pre}$, which is to be calibrated, because the intermediate features are better clustered and better aligned with GNN predictions \cite{kipf2016semi}. Hereafter we denote the intermediate feature map from the $l$-th layer of $G_{pre}$ as $X_{pre}^{(l)}$.

A naive solution would be to feed $X_{pre}^{(l)}$ directly into another GNN $G_{feat}$ for feature processing, in the hope that $G_{feat}$ implicitly takes feature similarity into account and produces nodewise temperatures $T$:
\begin{equation}
    T=G_{feat}(X^{(l)}_{pre}, \mathbf{A})\in \mathbb{R}^{|V|}.
\end{equation}

Unfortunately, this renders the number of parameters for $G_{feat}$ highly dependent on the number of $X_{pre}^{(l)}$'s dimensions. Specifically, if, for constants $h\ ,h' \in \mathbb{N}$, the first layer of $G_{feat}$ projects $X_{pre}^{(l)}$ from $\mathbb{R}^h$ to $\mathbb{R}^{h'}$, then $G_{feat}$ will contain at least $h \times h'$ parameters, which can easily result in overfitting when $h$ gets large. Thus, we desire a feature similarity mechanism that does not rely on input feature dimension or feature semantics.

Motivated by prototypical learning \cite{nassar2023protocon, snell2017prototypical}, for each class $k$, we first take the average feature as a classwise template,
\begin{equation}
    \forall k \in \mathcal{Y},\ \hat{\mu}_k := \frac{1}{|\mathcal{L}_k|}\mathop\sum_{i:v_i\in\mathcal{L}_k} x_i^{(l)},
\end{equation}
where $\mathcal{L}_k:=\{v_i\in \mathcal{L}| y_i=k \}$, and $x_i^{(l)}$ is the intermediate feature for $v_i$ in $X_{pre}^{(l)}$. Then, we define the similarity between $x_i^{(l)}$ and templates with the assistance of a distance measure $d(\cdot, \cdot)$:
\begin{equation}
    s_i := {\tt sim}(x_i^{(l)}, \{\hat{\mu}_k\}) := \zeta(\{d(x_i^{(l)}, \hat{\mu}_k)\}),
\end{equation}
where $\zeta(x)=\frac{x}{||x||_2}$ normalizes the input vector. Naturally, we consider $s_i$, the feature similarity between $x_i^{(l)}$ and template features, as a proxy of the similarity between $x_i^{(l)}$ and intermediate features of all the nodes from a class. Following Lee et al.\shortcite{lee2018simple}, we first compute variance matrix,
\begin{equation}
    \hat{\Sigma} := \frac{1}{| \mathcal{L} |} \mathop\sum_k \mathop\sum_{i\in \mathcal{L}_k} (x_i^{(l)}-\hat{\mu}_k)(x_i^{(l)}-\hat{\mu}_k)^T,
\end{equation}
and then induce the Mahalanobis distance \cite{mahalanobis2018generalized} accordingly:
\begin{equation}
    d(x_i^{(l)}, \hat{\mu}_k) := (x_i^{(l)} - \hat{\mu}_k)^T \hat{\Sigma}^{-1} (x_i^{(l)} - \hat{\mu}_k).
\end{equation}

Finally, we feed $\{ s_i \}$ and $\mathbf{A}$ to $G_{feat}$ to propagate feature-level similarities along the graph structure, i.e.,
\begin{equation}
    T_{feat} := G_{feat}(\{ s_i \}, \mathbf{A})\in \mathbb{R}^{|\mathcal{V}|}
\end{equation}
where $T_{feat}$ is the nodewise temperature estimate by feature-level similarity. 
\begin{figure}[t]
    \includegraphics[width=8cm]{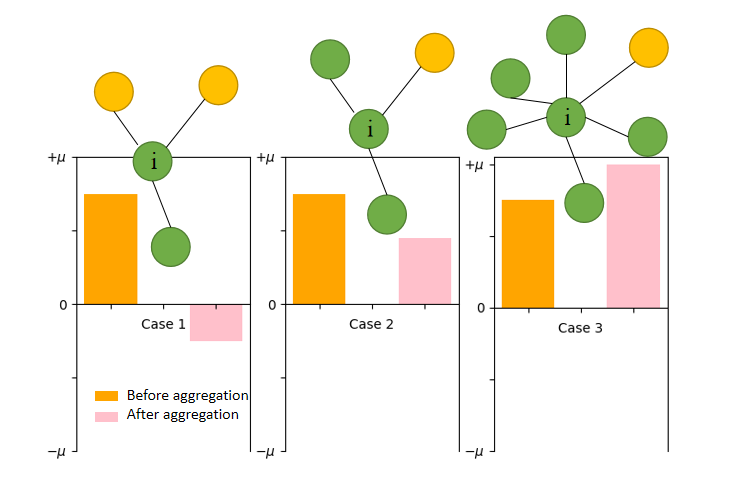}
    \caption{Illustrating nodewise representation movement cases. In the figure, we color nodes from different classes differently, and depict feature change of the central node i after a round of message passing in each case. The green and yellow classes' features are distributed around $+\mu$ and $-\mu$ respectively.}
    \label{fig:move_sim}
\end{figure}
\subsection{Representation movement similarity}
Recently, Yan et al.\shortcite{yan2022two} quantify nodewise representation movement dynamics of GNNs in node classification task, whose three cases of representation movement serve as the foundation of our representation movement similarity-aware mechanism. Hereafter, we provide the necessary backgrounds.

We denote the node degree of $v_i$ as $d_i$, and the homophily of $v_i$ is defined as: $h_i:=\mathbb{P}(y_i=y_j | v_j \in \mathcal{N}_i)$, in which $\mathcal{N}_i$ is $v_i$'s neighbor set. Finally, the expected relative degree of node $v_i$ is $\overline{r_i}:=\mathbb{E}_{\mathbf{A}|d_i} (\frac{1}{d_i} \mathop\sum_{j\in \mathcal{N}_i} r_{ij}|d_i)$, where $r_{ij}:=\sqrt{\frac{d_i + 1}{d_j + 1}}$. Then, Yan claims that node representation dynamics can be grouped as three cases:
\begin{itemize}
    \item Case 1: when $h_i$ is low, node representations move closer to the representations of the other class, whatever value $\overline{r_i}$ takes.
    \item Case 2: when $h_i$ is high but relative degree $\overline{r_i}$ is low, node representations still move closer to the other class but not as much as in the first case.
    \item Case 3: only when both $h_i$ and $\overline{r_i}$ are high, node representations tend to move away from the other class.
\end{itemize}

We provide Fig.~\ref{fig:move_sim} for an illustration. The existence of such dynamics can easily obscure information of logits and features, e.g. $v_i$ of case 1 from class 1 may end up having its logits similar to that of $v_j$, which is of case 2 from another class. Therefore, we desire our GNN calibrator to be able to decompose effects from similar nodewise representation movement behavior and nodewise input information, producing better calibrated results. Nonetheless, one difficulty of applying the theorem is the absence of ground-truth labels during training, making $h_i$ unaccessible. Worse still, when applied to GNNs, both $h_i$ and $\overline{r_i}$ are unable to differentiate messages from different neighbors, limiting the calibrator's model capacity. To circumvent these problems, we approximate homophily by $\hat{z}_i\cdot\hat{z}_j$. The relative degree information is considered by $\frac{d_i + 1}{d_j + 1}$. Furthermore, aware of correlation between ECE and distances to training nodes \cite{teixeira2019graph}, we estimate nodewise homophily and apply it to graph attention:
\begin{equation}
    \alpha_{i,j} := \mathop{{\tt softmax}}_{j\in \mathcal{N}_i} (\sigma (\frac{1}{\eta_i \eta_j} \hat{z}_i \ \cdot \ \hat{z}_j)),
\end{equation}
where $\eta_i$ is the distance from $v_i$ to the nearest training node, and $\sigma:=LeakyReLU$ \cite{xu2015empirical}. Then messages from neighbors are weighted and summed:
\begin{equation}
\label{eq:move}
    T_{move} = {\tt softplus}(\mathop\sum_{j \in \mathcal{N}_i} \alpha_{i,j} \eta_j (\frac{d_i + 1}{d_j + 1})^t \tilde{z}_j^T W),
\end{equation}
where $t$ is a hyperparameter, $\tilde{z}$ sorts the logits \cite{rahimi2020intra} and $W \in \mathbb{R}^{|\mathcal{Y}|}$ is a trainable parameter modeling the message from $v_j$. 

It is worth mentioning that representation movement dynamics was initially proven to relate to the oversmoothing problem\cite{yan2022two}, which refers to the problem that node features of GNNs converge towards the same values with the increase of model depth \cite{rusch2023survey}. Thus, the proceeding mechanism implies a relationship between oversmoothing and GNN calibration.

\subsection{Our model \& Calibration properties}
We formalize our GNN calibrator, SimCalib, as:
\begin{equation}
\begin{split}
    \forall v_i \in \mathcal{V},\ \hat{p}_i' & = \omega \cdot {\tt softmax}(\frac{\hat{z}_i}{T_{feat}}) \\ &\ \  + (1 - \omega) \cdot {\tt softmax}(\frac{\hat{z}_i}{T_{move}})
\end{split}
\end{equation}
where $\omega \in (0, 1)$ is a hyperparameter balancing feature and representation movement similarities. It is obvious that SimCalib is the composition of order-preserving functions and thus accuracy-preserving. We provide Fig.~\ref{fig:model} for illustration.
\section{Experiments}
In this section, we empirically demonstrate the effectiveness of our proposed method and evaluate the effects of various network designs.
\subsection{Experimental Setup}
In the experiments, we apply the commonly used equal-width binning scheme from Guo et al.\shortcite{guo2017calibration}: for any node subset $\mathcal{N} \subset \mathcal{V}$, samples are regrouped into $M$ equally spaced intervals according to their confidences, formally, $B_m := \{ v_i\in \mathcal{N} | \frac{m-1}{M} < \hat{c}_i \leq \frac{m}{M} \}$, to compute the expected calibration error (ECE) of the GNN:
\begin{equation}\label{eq:ece_def}
    {\tt ECE}=\mathop{\sum}^M_{m=1} \frac{|B_m|}{|\mathcal{N}|} |{\tt acc}(B_m) - {\tt conf}(B_m)|,
\end{equation}
where $acc(B_m)$ and $conf(B_m)$ are defined as:
\begin{equation}\label{eq:}
\begin{split}
{\tt acc}(B_m) & = \frac{1}{|B_m|} \mathop\sum_{i:v_i\in B_m} \mathbf{1}(y_i=\hat{y}_i), \\
{\tt conf}(B_m) & = \frac{1}{|B_m|} \mathop\sum_{i:v_i\in B_m} \hat{c}_i.
\end{split}
\end{equation}

To make a fair comparison, the evaluation protocol is mainly adopted from GATS. Specifically, we first train a series of GCNs \cite{kipf2016semi} and GATs \cite{velivckovic2017graph} with node classification on eight widely used datasets: Cora \cite{mccallum2000automating}, Citeseer \cite{giles1998citeseer}, Pubmed \cite{sen2008collective}, CoraFull \cite{bojchevski2017deep}, and the four Amazon datasets \cite{shchur2018pitfalls}. Then we train calibrators on top of the pretrained nodewise classifiers to evaluate its calibration performance. After training, we evaluate models by ECE with $M=15$ equally sized bins. To reduce the influence of randomness, we randomly assign 15\% of nodes as $\mathcal{L}$, and the rest as $\mathcal{U}$, and we repeat this assignment process with randomness five times for each dataset. Once $\mathcal{L}$ has been sampled, we use three-fold cross-validation on it. Also, in each fold we randomly initialize our models five times. Therefore, this results in a total of 75 runs for experiment, the mean and standard deviation of which are finally reported. Full implementation details are presented in the Appendix.

    \begin{table*}[t]
        \centering
        \begin{tabular}{ c|c|cccc|ccc } 
             \toprule[2pt]
              Dataset & Backbone & UnCal & TS & VS & ETS & CaGCN & GATS & SimCalib \\
             \hline
              \multirow{2}{*}{Cora} & GCN & 13.04$\pm$5.22 & 3.92$\pm$1.29 & 4.36$\pm$1.34 & 3.79$\pm$3.54 & 5.29$\pm$1.47 & 3.64$\pm$1.34 & \textbf{3.32$\pm$0.99} \\
              & GAT & 23.31$\pm$1.81 & 3.69$\pm$0.90 & 3.30$\pm$1.12 & 3.54$\pm$1.01 & 4.09$\pm$1.06 &  3.18$\pm$0.90 &  \textbf{2.90$\pm$0.87}\\
             \hline
              \multirow{2}{*}{Citeseer} & GCN & 10.66$\pm$5.92 & 5.15$\pm$1.50 & 4.92$\pm$1.44 & 4.65$\pm$1.69 & 6.86$\pm$1.41 & 4.43$\pm$1.30 & \textbf{3.94$\pm$1.12}\\
              & GAT & 22.88$\pm$3.53 & 4.74$\pm$1.47 & 4.25$\pm$1.48 & 4.11$\pm$1.64 & 5.75$\pm$1.31  & \textbf{3.86$\pm$1.56}  & 3.95$\pm$1.30 \\
             \hline
              \multirow{2}{*}{Pubmed} & GCN & 7.18$\pm$1.51 & 1.26$\pm$0.28 & 1.46$\pm$0.29 & 1.24$\pm$0.30 & 1.09$\pm$0.52 & 0.98$\pm$0.30 & \textbf{0.93$\pm$0.32} \\
              & GAT & 12.32$\pm$0.80 & 1.19$\pm$0.36 & 1.00$\pm$0.32 & 1.20$\pm$0.32 & 0.98$\pm$0.31 &  1.03$\pm$0.32 & \textbf{0.95$\pm$0.35}\\
             \hline
              \multirow{2}{*}{Computers} & GCN & 3.00$\pm$0.80 & 2.65$\pm$0.57 & 2.70$\pm$0.63 & 2.58$\pm$0.70 & 1.72$\pm$0.53 & 2.23$\pm$0.49 & \textbf{1.37$\pm$0.33}\\
              & GAT & 1.88$\pm$0.82 & 1.63$\pm$0.46 & 1.67$\pm$0.52 & 1.54$\pm$0.67 & 2.03$\pm$0.80 &  1.39$\pm$0.39 & \textbf{1.08$\pm$0.33}\\
             \hline
              \multirow{2}{*}{Photo} & GCN & 2.24$\pm$1.03 & 1.68$\pm$0.63 & 1.75$\pm$0.63 & 1.68$\pm$0.89 & 1.99$\pm$0.56 & 1.51$\pm$0.52 & \textbf{1.36$\pm$0.59}\\
              & GAT & 2.02$\pm$1.11 & 1.61$\pm$0.63 & 1.63$\pm$0.69 & 1.67$\pm$0.73 & 2.10$\pm$0.78 &  1.48$\pm$0.61 & \textbf{1.29$\pm$0.55}\\
             \hline
              \multirow{2}{*}{CS} & GCN & 1.65$\pm$0.92 & 0.98$\pm$0.27 & 0.96$\pm$0.30 & 0.94$\pm$0.24 & 2.27$\pm$1.07 & 0.88$\pm$0.30 & \textbf{0.81$\pm$0.30}\\
              & GAT & 1.40$\pm$1.25 & 0.93$\pm$0.34 & 0.87$\pm$0.35 & 0.88$\pm$0.33 & 2.52$\pm$1.04 &  \textbf{0.81$\pm$0.30} & 0.83$\pm$0.32\\
             \hline
              \multirow{2}{*}{Physics} & GCN & 0.52$\pm$0.29 & 0.51$\pm$0.19 & 0.48$\pm$0.16 & 0.52$\pm$0.19 & 0.94$\pm$0.51 & 0.46$\pm$0.16 & \textbf{0.39$\pm$0.14}\\
              & GAT & 0.45$\pm$0.21 & 0.50$\pm$0.21 & 0.52$\pm$0.20 & 0.50$\pm$0.21 & 1.17$\pm$0.42 & 0.42 $\pm$0.14 & \textbf{0.40$\pm$0.13}\\
             \hline
              \multirow{2}{*}{CoraFull} & GCN & 6.50$\pm$1.26 & 5.54$\pm$0.43 & 5.76$\pm$0.42 & 5.38$\pm$0.49 & 5.86$\pm$2.52 & 3.76$\pm$0.74 & \textbf{3.22$\pm$0.74}\\
              & GAT & 4.73$\pm$1.39 & 4.00$\pm$0.50 & 4.17$\pm$0.43 & 3.89$\pm$0.56 & 6.55$\pm$3.69 & 3.54$\pm$0.63 & \textbf{3.40$\pm$0.91}\\
             \bottomrule[2pt]
        \end{tabular}
        \caption{GNN calibration results of SimCalib and other baseline approaches in terms of ECE (\%), where lower is better. For each experiment, the best result is displayed in bold. UnCal stands for the uncalibrated backbones.}
        \label{table:main}
    \end{table*}

    \begin{table}[t]
        \small
        \centering
        \vspace{0em}
        \begin{tabular}{c|ccc } 
             \toprule[2pt]
              Backbone & Uncal & GATS & SimCalib \\
             \hline
              GCN & 16.23 & 15.80 & \textbf{15.44}\\
              GAT & 16.27 & 15.52 & \textbf{15.17}\\
             \bottomrule[2pt]
        \end{tabular}
        \caption{Mean and standard variation of ACE (\%) of SimCalib and baselines on CoraFull.}
        \label{table:ace}
        \vspace{0em}
    \end{table}
\subsection{Performance Comparison}
We benchmark SimCalib against a variety of baselines on GNN calibration tasks:
\begin{itemize}
    \item Temperature scaling(TS) applies a global temperature to scale every nodewise logits.
    \item Vector scaling(VS) scales and adds a bias to each class in a class-wise manner.
    \item Ensemble temperature scaling(ETS) softens probabilistic outputs by learning an ensemble of uncalibrated, TS-calibrated and uniform distribution.
    \item Graph convolution network as a calibration function (CaGCN) uses a GCN to process logits, producing nodewise temperatures.
    \item Graph attention temperature scaling (GATS) identifies factors that influence GNN calibration, and addresses them with graph attention mechanism.
\end{itemize}

We also report the ECEs for uncalibrated predictions as a reference. Among the baselines, TS, VS and ETS are designed for standard i.i.d. multi-class classification. CaGCN and GATS propagate information along the graph structure, and produce separate nodewise temperatures.

For all the experiments, the pretrained GNN classifiers will be frozen, and its first-layer feature map, together with the logits, will be fed into our calibration model as inputs. We train calibrators on validation sets by minimizing NLL loss, and validate it on the training set, following the common practice \cite{wang2021confident}. We provide details of comparison settings and hyperparameters in Appendix. The calibration results are summarized in Table \ref{table:main}. We also provide the comparisons on adaptive calibration error (ACE) in Table \ref{table:ace}.

Overall, SimCalib consistently produces well calibrated results for all the GNN backbones on every dataset. It sets a new SOTA for all experiments, with two exceptions of GAT on Citeseer(2nd best) and GAT on CS(2nd best), on which SimCalib performs worse than GATS by at most 3\%. In contrast, SimCalib's improvements are more statistically significant, reducing average ECE by 10.4\% compared to GATS.

With Wilcoxon signed test \cite{wilcoxon1992individual} backed by scipy \cite{2020SciPy-NMeth}, we claim that our model is superior to the previous SOTA model, namely, GATS, with confidence 99.9\% and $p=7.63\times 10^{-4}$.
    \begin{table*}[t]
        \centering
        \begin{tabular}{ c|c|cc|ccc|c } 
             \toprule[2pt]
              Dataset & Backbone & $\text{SimCalib}_l$ & $\text{SimCalib}_m$ & $\text{SimCalib}_h$ & $\text{SimCalib}_r$ & $\text{SimCalib}_n$ & SimCalib \\
             \hline
              \multirow{2}{*}{Cora} & GCN & 3.58$\pm$0.97 & 3.86$\pm$1.78 & 3.30$\pm$1.76 & 3.87 $\pm$ 1.77 & 4.16 $\pm$1.79 & 3.32$\pm$0.99 \\
               & GAT & 2.88$\pm$0.88 & 3.40$\pm$1.32 & 3.02$\pm$1.27 & 3.52 $\pm$1.26 & 3.68$\pm$1.39 &  2.90$\pm$0.87 \\
             \hline
              \multirow{2}{*}{Citeseer} & GCN & 4.24$\pm$1.61 & 4.57$\pm$1.92 & 4.36$\pm$1.84 & 4.75$\pm$1.89 & 4.96 $\pm$1.93 & 3.94$\pm$1.12 \\ 
              & GAT & 4.22$\pm$1.51 & 4.41$\pm$2.29 & 3.93$\pm$2.25 &  4.65$\pm$2.21 & 4.73$\pm$2.20 &  3.95$\pm$1.30 \\
             \hline
              \multirow{2}{*}{Photo} & GCN & 1.50$\pm$0.50 & 1.42$\pm$0.63 & 1.38$\pm$0.71 & 1.52 $\pm$0.65 & 1.58 $\pm$ 0.71 & 1.36$\pm$0.59\\
              & GAT & 1.39$\pm$0.58 &  1.44$\pm$0.55 & 1.36$\pm$0.54  &  1.55$\pm$0.54 & 1.52$\pm$0.59 &  1.29$\pm$0.55 \\
             \hline
              \multirow{2}{*}{CoraFull} & GCN & 3.47$\pm$0.79 & 3.91$\pm$0.79 & 3.76$\pm$0.82 & 3.18$\pm$0.73 & 3.33$\pm$0.90 & 3.22$\pm$0.74\\
              & GAT & 3.84$\pm$0.80 & 3.27$\pm$0.84 & 3.59$\pm$0.88 & 3.35$\pm$0.81 & 3.21$\pm$0.84 & 3.40$\pm$0.91 \\
             \bottomrule[2pt]
        \end{tabular}
        \caption{Ablation study results in terms of ECE (\%). Overall, all designs are critical and removing any of them results a general decrease in performance.}
        \label{table:ablation}
    \end{table*}
\begin{figure}[t]
\centering
\includegraphics[width=8cm]{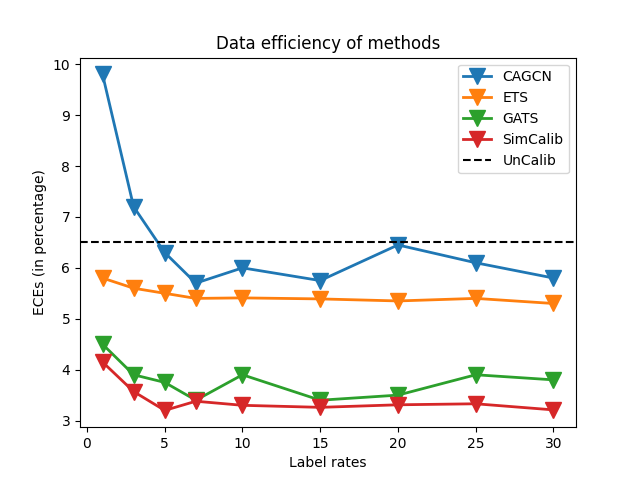}
\caption{ECEs (\%) of CaGCN, ETS, GATS and SimCalib, with different amounts of calibration data. For reference, we also plot the result for uncalibrated backbone as the dashed line.}
\label{fig:efficiency}
\end{figure}

\subsection{Correlation between feature similarity and calibration improvement}
Furthermore, we conduct experiments on CoraFull to assess the correlation between calibration improvement and nodewise similarity. We opt for CoraFull because it is the most complicated dataset of the eight, with 19,793 nodes, 126,842 edges, 70 classes and 8710 features, which makes it well representative of the real-world scenario.
\begin{table}[t]
        \small
        \centering
        \vspace{0em}
        \begin{tabular}{c|cccc } 
             \toprule[2pt]
              Backbone & w=0.5 & w=0.6 & w=0.8 & w=0.9 \\
             \hline
              GCN & 3.34$\pm$0.84 & 3.22$\pm$0.74 & 3.32$\pm$0.69 & 3.56$\pm$0.70 \\
              GAT & 3.52$\pm$1.02 & 3.40$\pm$0.91 & 3.46$\pm$0.79 & 3.68$\pm$0.66\\
             \bottomrule[2pt]
        \end{tabular}
        \caption{Mean and standard variation of ECE (\%) of SimCalib on CoraFull with different $w$.}
        \label{table:diff_w}
        \vspace{0em}
    \end{table}

\begin{figure}[t]
    \centering
    \includegraphics[width=8cm]{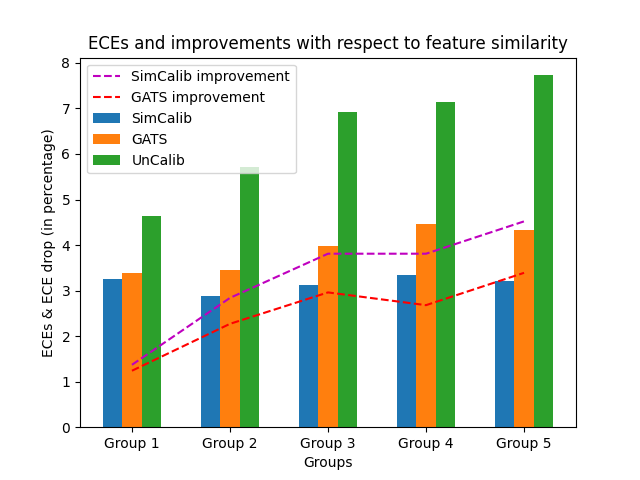}
    \caption{Figure of correlation between feature-level similarity and calibration improvement. We also illustrate the calibration improvement in terms of ECE (\%) with the dashed lines. The groups are sorted in an ascending order with respect to nodewise Mahalanobis distances to the nearest templates.}
    \label{fig:feat_sim}
\end{figure}

In Fig.~\ref{fig:feat_sim}, we visually examine the correlation between feature similarity and calibration improvement by comparing the calibration performance of GATS, an uncalibrated GNN backbone and SimCalib. We group nodes by the Mahalanobis distances to the nearest template representation, and display the group-level ECEs as bars and ECE improvements as dashed lines. The figure shows that the miscalibration issue worsens with the decrease of feature similarity, while our calibration strategy calibrates the nodewise confidence in a consistent way. We believe that with the decrease of feature-level similarity, the samples become more outlying in the feature space, indistinguishable from samples from other classes, and thus it gets harder for the uncalibrated model to accurately tune its confidence in congruence with ground-truth prediction accuracy. In contrast, our model successfully overcomes such ambiguity in feature space by taking feature-level similarity into account, thereby consistenly calibrating samples of various feature similarity with similar expected calibration error. Therefore, this results in stronger improvement for SimCalib when feature similarity gets weaker. Although we discover a similar pattern for GATS, its ECE improvement is weaker than SimCalib when features become dissimilar. We attribute the performance gap to the feature-similarity-aware mechanism. The observation aligns with our theoretical hypothesis in that it suggests that feature similarity indeed plays a critical role in GNN calibration. 
\subsection{Data-efficiency and expressivity of SimCalib}
In addition, we analyze the data-efficiency and expressivity of SimCalib for GNN calibration. For this, we reuse the GCN classifier pretrained on CoraFull, and compare the expected calibration errors between baselines and SimCalib, with different amounts of calibration data. The results are shown in Fig.~\ref{fig:efficiency}. From the figure, we draw that SimCalib is both data-efficient and expressive. SimCalib does not require a lot of labels to perform decently, consistenly outperforming all the baselines under all label rates. Moreover, SimCalib also expresses robustness to label rates.
\subsection{Ablation Study}
To understand the effects of the two similarity-oriented mechanisms, we conduct a thorough ablation study in this section. The results are shown in Table \ref{table:ablation} and Table \ref{table:diff_w}. Overall, each mechanism plays a critical role in GNN calibration and removing any will in general decrease performance while increasing variances.

\textbf{Effect of feature similarity}
In order to decompose the effects from number of trainable parameters and feature-similarity mechanism, we investigate the performance of two different models: SimCalib with only representation movement similarity($\text{SimCalib}_m$) and SimCalib with $s_i$ replaced by $G_{pre}$'s output logits($\text{SimCalib}_l$). Comparing $\text{SimCalib}_m$ with $\text{SimCalib}_l$, we see that the calibration performance slightly improves with more trainable parameters. However, the improvements are rather moderate, unable to match the performance of SimCalib. We hypothesize that since logits information has already been integrated into the calibration process in the nodewise representation movement similarity, adding an extra branch of logits propagation only helps GNN calibration by introducing more parameters.

\textbf{Effect of representation movement similarity}
Our representation movement similarity mechanism consists of two aspects of network designs, i.e. homophily term $\hat{z}_i \cdot \hat{z}_j$ and relative degree $(\frac{d_j + 1}{d_i + 1})^t$, therefore we design three models to analyze the effects of various components in the representation movement similarity mechanism. Particularly, we test the calibration performance of $\text{SimCalib}_h$ that disables relative degree, $\text{SimCalib}_r$ that disables homophily, and $\text{SimCalib}_n$ in which neither takes effects. We can easily draw from the experiments that the integrity of representation movement similarity mechanism is important to calibration performance and removing of any results in a worsened GNN calibrator. We attribute the performance drop to the inability of the calibrator to decompose effects from nodewise input information and effects from representation movement.

\section{Conclusions}
In this work, we provide theoretical analysis on the graph calibraion problem and prove that nodewise similarity plays an important role in the solution. We consider feature and nodewise representation movement similarities, which are quantified by Gaussian-induced Mahalanobis distances and homophily \& relative degrees, respectively. Based on the mechanisms, we propose a novel calibrator, SimCalib, tailored for GNN calibration. SimCalib is data-efficient, expressive and accuracy-preserving at the same time. Our extensive experiments demonstrate the effectiveness of SimCalib by achieving state-of-the-art performances for GNN calibration on various datasets and for different backbones. Moreover, our experiments exhibit a correlational relationship between nodewise similarity and calibration improvement, in alignment with our theoretical results. Our work has the potential to be employed in cost-sensitive scenarios. Additionally our work is the first to reveal a non-trivial relationship between oversmoothing and GNN calibration problems.

\section{Acknowledgement}
This work is supported by National Natural Science Foundation of China (62076144, 62306260), Shenzhen Science and Technology Program (WDZC20220816140515001, JCYJ20220818101014030), and the Center for Perceptual and Interactive Intelligence (CPII) Ltd under the Innovation and Technology Commission’s InnoHK Scheme.
\bibliography{aaai24}

\newpage
\clearpage

\appendix
\section{Appendix}
\subsection{Proofs}
In this section, we illustrate our mathematic derivation of Thearom 1. The proof may seem horiffying, but the idea is simple, which is to compare the ECE of joint optimization and that of separated optimization based on probability concentation inequalities (of Gaussian and Chi-square variables). Then, 

\begin{proposition}\label{prop}
    The learned model predicts $\hat{\vy}_i=1$ whenever $\hat{\vmu}_i\vx_i\geq0$.
\end{proposition}
\begin{proof} 
    According to expected cost minimum (ECM) with equal mis-classification costs, 
    the learned model predicts $\hat{\vy}_i=1$ if ${\tt p}(\vx_i|\hat{\vmu}_i,\sigma_i\mI)\geq{\tt p}(\vx_i|-\hat{\vmu}_i,\sigma_i\mI)$, i.e., $\hat{\vmu}_i\vx_i\geq0$. 
\end{proof}
\begin{lemma}
\label{lem:ece}
\label{prop:gaussian_ece}
    For the learned model, the ECE measure is 
\begin{eqnarray}
    {\tt ECE} &=& \sE_{v=\hat{\vmu}^\top\vx}\Bigg[\Bigg|\frac{1}{e^{-\frac{2\hat{\vmu}^\top\vmu}{||\hat{\vmu}||^2}v}+1}-\frac{1}{e^{-2v}+1}\Bigg|\Bigg]
\end{eqnarray}
\end{lemma}
\begin{proof}
Since $\tt\hat{p}(Y=1|\vx;\hat{\vmu})=\frac{1}{e^{-2\hat{\vmu}^\top\vx}+1}$,
    \begin{eqnarray}
        &\sE_p[Y=1|\tt\hat{p}(Y=1|\vx;\hat{\vmu})=p] \\
        = &\sE_{v=\hat{\vmu}^\top\vx}[Y=1|\tt\hat{p}(Y=1|\vx;\hat{\vmu})=\frac{1}{e^{-2v+1}}] &\\
        = &\sE_{v=\hat{\vmu}^\top\vx}[Y=1|\hat{\vmu}^\top\vx=v] &\\
        = &\frac{\ttp(\hat{\vmu}^\top\vx=v|Y=1)}{\ttp(\hat{\vmu}^\top\vx=v|Y=1)+\ttp(\hat{\vmu}^\top\vx=v|Y=-1)}& \\
        = &\frac{1}{e^{-\frac{2\hat{\vmu}^\top\vmu}{||\hat{\vmu}||^2}v}+1} &
    \end{eqnarray}
thus,
    \begin{eqnarray}
    {\tt ECE} &=& \sE_{p}\Bigg[\Bigg|\sE[Y=1|\tt\hat{p}(Y=1|\vx;\hat{\vmu})=p]-p\Bigg|\Bigg] \\
    &=&\sE_{v=\hat{\vmu}^\top\vx}\Bigg[\Bigg|\sE[Y=1|\tt\hat{p}(Y=1|\vx)=\frac{1}{e^{-2v}+1}] \nonumber\\
    &-&\frac{1}{e^{-2v}+1}\Bigg|\Bigg] \nonumber\\
    &=&\sE_{v=\hat{\vmu}^\top\vx}\Bigg[\Bigg|\frac{1}{e^{-\frac{2\hat{\vmu}^\top\vmu}{||\hat{\vmu}||^2}v}+1}-\frac{1}{e^{-2v}+1}\Bigg|\Bigg]
\end{eqnarray}
\end{proof}
where $\tt\hat{p}$ is the probability estimation from the learned model.

\begin{lemma}
\label{lem:up_bound}
    There exists numerical constant $c_1$, when $d/n$ is sufficiently large, $\frac{\hat{\vmu}^\top\vmu}{||\hat{\vmu}||^2}\leq 1$, with high probability,
    \begin{eqnarray}
    \ttp\Big(\frac{\hat{\vmu}^\top\vmu}{||\hat{\vmu}||^2}\leq1\Big)\geq 1-e^{c_1d/32}
    \end{eqnarray}
\end{lemma}
\begin{proof}
    Let $\veps_i=\bar{\vmu}_i-\vmu_i$, $\veps_j=\bar{\vmu}_j-\vmu_j$, then $\veps_i\sim\mathcal{N}(0,\frac{\sigma^2}{n}\mI)$, $\veps_j=\sim\mathcal{N}(0,\frac{\sigma^2}{n}\mI)$. Let $\vdelta=\frac{1}{1+a^2}\veps_i+\frac{a}{1+a^2}\veps_j\sim\mathcal{N}(0,\frac{\sigma^2}{(1+a^2)n})$, from Eq.~\ref{eq:original_estimator}, we have $\hat{\vmu}_i=\vmu_i + \frac{1}{1+a^2}\veps_i + \frac{a}{1+a^2}\veps_j=\vmu_i+\vdelta$, and $||\vdelta||^2\sim \frac{\sigma^2}{(1+a^2)n}\chi_d^2$. 
    
    Given $\frac{\hat{\vmu}^\top\vmu}{||\hat{\vmu}||^2}=\frac{||\vmu||^2 + \vmu^\top\vdelta}{||\vmu||^2 + ||\vdelta||^2+ 2\vmu^\top\vdelta}=\frac{1}{2}+\frac{1}{2}\frac{||\vmu||^2 - ||\vdelta||^2}{||\vmu||^2 + ||\vdelta||^2+ 2\vmu^\top\vdelta}$, 
    according to concentration inequalities $\ttp(||\vdelta||^2\leq \frac{d\sigma^2}{2(1+a^2)n})\leq e^{-d/16},\ttp(\vmu^\top\vdelta\leq-\frac{d\sigma}{4(1+a^2)^{1/2}n^{1/2}})\leq e^{-d/32}$, 
    we have
\begin{eqnarray}
    &\ttp\Bigg(\frac{\hat{\vmu}^\top\vmu}{||\hat{\vmu}||^2}\geq \frac{1}{2}+\frac{1}{2}\frac{d-\frac{d\sigma^2}{2(1+a^2)n}}{d + \frac{d\sigma^2}{2(1+a^2)n} -2 \frac{d\sigma}{4(1+a^2)^{1/2}n^{1/2}} }\Bigg) \nonumber\\
    \leq & \ttp(||\vdelta||^2\leq \frac{d\sigma^2}{2(1+a^2)n}) + \ttp(\vmu^\top\vdelta\leq-\frac{d\sigma}{4(1+a^2)^{1/2}n^{1/2}}) & \nonumber\\
    \leq &2e^{-d/16}+e^{-d/32} & \nonumber\\
    \leq  & e^{-c_1d/32} \nonumber
\end{eqnarray}
thus,
\begin{eqnarray}
    \ttp(\frac{\hat{\vmu}^\top\vmu}{||\hat{\vmu}||^2}\leq 1)  \geq  1-e^{-c_1d/32}    
\end{eqnarray}
with 
$(\frac{d}{n})^{1/2}\geq\frac{1+a^2}{4}$.
\end{proof}
\begin{lemma}
\label{lem:low_bound}
    With sufficiently large $d/n$,  $\frac{\bar{\vmu}^\top\vmu}{||\bar{\vmu}||^2}\geq 1/2$, with high probability,
\begin{eqnarray}
    \ttp(\frac{\bar{\vmu}^\top\vmu}{||\bar{\vmu}||^2}\geq 1/2)\geq 1-3e^{d/8}
\end{eqnarray}
\end{lemma}
\begin{proof}
    Given $\frac{\bar{\vmu}^\top\vmu}{||\bar{\vmu}||^2}=\frac{||\vmu||^2 + \vmu^\top\veps}{||\vmu||^2 + ||\veps||^2+ 2\vmu^\top\veps} = \frac{1}{2}+\frac{1}{2}\frac{||\vmu||^2 - ||\veps||^2}{||\vmu||^2 + ||\veps||^2+ 2\vmu^\top\veps}$, $||\veps||^2\sim \frac{\sigma^2}{n}\chi_d^2$,
    according to concentration inequalities $ \ttp(||\veps||^2\leq\frac{d\sigma^2}{4n})\leq e^{-d/8},\ttp(||\veps||^2\geq\frac{2d\sigma^2}{n})\leq e^{-d/8}$, $\ttp(\vmu^\top\veps\geq\frac{d\sigma}{2n^{1/2}})\leq e^{-d/8},\ttp(\vmu^\top\veps\leq-\frac{d\sigma}{2n^{1/2}})\leq e^{-d/8}$, we have
    \begin{eqnarray}
        &\ttp(\frac{\bar{\vmu}^\top\vmu}{||\bar{\vmu}||^2}\leq \frac{1}{2}+\frac{1}{2}\frac{d-\frac{2d\sigma^2}{n}}{ d+\frac{d\sigma^2}{4n} - 2\frac{d\sigma}{2n^{1/2}}} ) \nonumber \\
        \leq & \ttp(||\veps||^2\geq \frac{2d\sigma^2}{n}) + \ttp(||\veps||^2\leq \frac{d\sigma^2}{4n}) + \ttp(\vmu^\top\veps\leq-\frac{d\sigma}{2n^{1/2}}) & \nonumber \\
        \leq & 3e^{-d/8} \nonumber
    \end{eqnarray}
hence, 
\begin{eqnarray}
    \ttp\Big(\frac{\bar{\vmu}^\top\vmu}{||\bar{\vmu}||^2}\geq \frac{1}{2}\Big) \geq 1-3e^{-d/8}
\end{eqnarray}
with sufficiently large $(\frac{d}{n})^{1/2} \geq n$. 
\end{proof}

\begin{lemma}
\label{lem:mu_bar}
With the parameter setting and sufficiently large $d/n$,
\begin{eqnarray}
    \ttp\Bigg(\frac{\bar{\vmu}^\top\vmu}{||\bar{\vmu}||^2}\leq \Bigg(2+\frac{\frac{d\sigma^2}{4n}-d}{d+\frac{d\sigma}{2n^{1/2}}}\Bigg)^{-1}\Bigg)\geq 1-2e^{-d/8} 
\end{eqnarray}
\end{lemma}
\begin{proof}
    $\frac{||\bar{\vmu}||^2}{\bar{\vmu}^\top\vmu}=\frac{||\vmu||^2 + ||\veps||^2+ 2\vmu^\top\veps}{||\vmu||^2 + \vmu^\top\veps}=2 + \frac{||\veps||^2-||\vmu||^2}{||\vmu||^2 + \vmu^\top\veps}$.

\begin{eqnarray}
    &\ttp(\frac{||\bar{\vmu}||^2}{\bar{\vmu}^\top\vmu} \leq 2+\frac{\frac{d\sigma^2}{4n}-d}{d+\frac{d\sigma}{2n^{1/2}}}) \\
    \leq & \ttp(||\veps||^2\leq\frac{d\sigma^2}{4n}) + \ttp(\vmu^\top\veps\geq \frac{d\sigma}{2n^{1/2}})\\
    \leq & 2e^{-d/8}
\end{eqnarray}
Hence,
\begin{eqnarray}
    \frac{\bar{\vmu}^\top\vmu}{||\bar{\vmu}||^2}\leq \Bigg(2+\frac{\frac{d\sigma^2}{4n}-d}{d+\frac{d\sigma}{2n^{1/2}}}\Bigg)^{-1}
\end{eqnarray}
with probability at least $1-2e^{-d/8}$.
\end{proof}

\begin{lemma}
\label{lem:mu_hat}
With sufficiently large $d/n$ and $a^2$, 
\begin{eqnarray}
    \ttp\Bigg(\frac{\hat{\vmu}^\top\vmu}{||\hat{\vmu}||^2}\geq \Bigg(2+\frac{\frac{d\sigma^2}{4n}-d}{d+\frac{d\sigma}{2n^{1/2}}}\Bigg)^{-1}\Bigg) \geq 1-2e^{-d/8} 
\end{eqnarray}
\end{lemma}
\begin{proof}
$\frac{||\hat{\vmu}||^2}{\hat{\vmu}^\top\vmu}=\frac{||\vmu||^2 + ||\vdelta||^2+ 2\vmu^\top\vdelta}{||\vmu||^2 + \vmu^\top\vdelta}=2 + \frac{||\vdelta||^2-||\vmu||^2}{||\vmu||^2 + \vmu^\top\vdelta}$

\begin{eqnarray}
    &\ttp\Big(\frac{||\hat{\vmu}||^2}{\hat{\vmu}^\top\vmu} \geq 2+\frac{\frac{2d\sigma^2}{(1+a^2)n}-d}{d-\frac{d\sigma}{2(1+a^2)^{1/2}n^{1/2}}}\Big) \\
    \leq &\ttp(||\vdelta||^2\geq\frac{2d\sigma^2}{(1+a^2)n}) + \ttp(\vmu^\top\vdelta\leq\frac{-d\sigma}{2(1+a^2)^{1/2}n^{1/2}}) \\
    \leq & 2e^{-d/8}
\end{eqnarray}
with $(1+a^2)^{1/2} \geq \frac{t^2-4+\sqrt{16+256t+248t^2+64t^3+t^4}}{4t+8}\geq t/2,t=(d/n)^{1/4}$, we have $\frac{\frac{2d\sigma^2}{(1+a^2)n}-d}{d-\frac{d\sigma}{2(1+a^2)^{1/2}n^{1/2}}} \leq \frac{\frac{d\sigma^2}{4n}-d}{d+\frac{d\sigma}{2n^{1/2}}}$
Hence, with sufficient large $a^2$, i.e., large linear correlation between nodes,  
\begin{eqnarray}
    \ttp\Bigg(\frac{\hat{\vmu}^\top\vmu}{||\hat{\vmu}||^2}\geq \Bigg(2+\frac{\frac{d\sigma^2}{4n}-d}{d+\frac{d\sigma}{2n^{1/2}}}\Bigg)^{-1}\Bigg) \geq 1-2e^{-d/8}.
\end{eqnarray}

\end{proof}

\begin{thm}
 Under the above parameter setting, there exist numerical constants $c_0,c_2$, with $d/n>c_0$ and $a^2>(\frac{d}{n})^{1/4}/2$,
\begin{eqnarray}
    {\tt ECE}_{\hat\vmu} \leq {\tt ECE}_{\bar\vmu} ~~\text{with probability}~\geq 1-e^{-c_2d/32}. 
\end{eqnarray}    
\end{thm}
\begin{proof}
    According to Lemma~\ref{lem:up_bound}, with sufficiently large $(\frac{d}{n})^{1/2}\geq \frac{1+a^2}{4}$, there exists numerical constant $c_1$, such that
    \[
    \ttp\Big(\frac{\hat{\vmu}^\top\vmu}{||\hat{\vmu}||^2}\leq 1\Big)  \geq  1-e^{-c_1d/32}. 
    \]
    
    According to Lemma~\ref{lem:low_bound}, with sufficiently large $(\frac{d}{n})^{1/2}\geq n$,
    \[
    \ttp\Big(\frac{\bar{\vmu}^\top\vmu}{||\bar{\vmu}||^2}\geq \frac{1}{2}\Big) \geq 1-3e^{-d/8}
    \]

    From Lemma~\ref{lem:mu_bar} and \ref{lem:mu_hat}, with sufficiently large $a^2>(\frac{d}{n})^{1/4}/2$, 
    \[
    \ttp\Bigg(\frac{\hat{\vmu}^\top\vmu}{||\hat{\vmu}||^2}\geq \frac{\bar{\vmu}^\top\vmu}{||\bar{\vmu}||^2}\Bigg) \geq 1-4e^{-d/8}
    \]
    hence, with sufficiently large $(\frac{d}{n})^{1/2}\geq c_0$, 
    \[
    \ttp\Bigg(1\geq\frac{\hat{\vmu}^\top\vmu}{||\hat{\vmu}||^2}\geq \frac{\bar{\vmu}^\top\vmu}{||\bar{\vmu}||^2}\geq\frac{1}{2}\Bigg) \geq 1-e^{-c_1d/32}-7e^{-d/8},
    \]
    and $\frac{1}{e^{-\frac{2\hat{\vmu}^\top\vmu}{||\hat{\vmu}||^2}v}+1}$ is always closer to $\frac{1}{e^{-2v}+1}$ than $\frac{1}{e^{-\frac{2\bar{\vmu}^\top\vmu}{||\bar{\vmu}||^2}v}+1}$ with various $v$. Thus, 
    \[
    \ttp({\tt ECE}_{\hat\vmu} \leq {\tt ECE}_{\bar\vmu})\geq 1-e^{-c_2d/32}.
    \]
\end{proof}

According to Theorem 1, considering the high correlation by jointly optimizing likelihoods of correlated nodes, leads to lower ECE than separately optimizing likelihoods, with high probability $\geq 1-e^{-c_2d/32}$. For example, given 4 graphs ($n=4$)  with 64-dim ($d=64$) features and linear correlation coefficients $|a|=2.66$, the probability $\geq 0.84$.

\subsection{Implementation Details}
Throughout all the experiments, we fix the global random seed to be 10, remaining the same with GATS. The seed eliminates randomness from python, numpy, pytorch and cuda. Our experiments are run on an Ubuntu 20.04 operating system, with a Nvidia V100 GPU, 64GB RAM and i9-13900K CPU. We mainly base our code on pytorch \cite{paszke2019pytorch} 1.12.1 and torch geometric 2.0.1. In all the experiments, $G_{feat}$ is a GNN or GAT while $G_{move}$ implements equation\ref{eq:move} with at most 8 heads, although we find 1 or 2 heads are sufficient for most experiments. For each experiment, we do a small grid search on validation set to determine $w$ and $t$, which can take values from $\{0.6, 0.8, 0.9\}$ and $\{0.3, 0.5, 1.0\}$ respectively. Following GATS, we train pretrained classifiers on Cora, Citeseer and Pubmed with a weight decay of 5e-4, and none on other datasets. We also conduct a coarse grid search to identify learning rate and number of heads on each dataset. All the hyperparameters are provided as a config file in our code appendix for reproducibility. 

We train GATS and other baselines with the open-sourced code of GATS, and find that the outcomes do not exhibit any statistically significant difference with the results reported by GATS. Thus we adopt the reported performance from GATS as our baselines. It is worth mentioning that our network consumes twice the amount of trainable parameters as GATS, but we cannot report the performance of GATS with the same number of parameters because GATS gets its performance declined with twice the number of attention heads \cite{teixeira2019graph}.

\subsection{Data efficiency of SimCalib}
\begin{figure}
    \includegraphics[width=8cm]{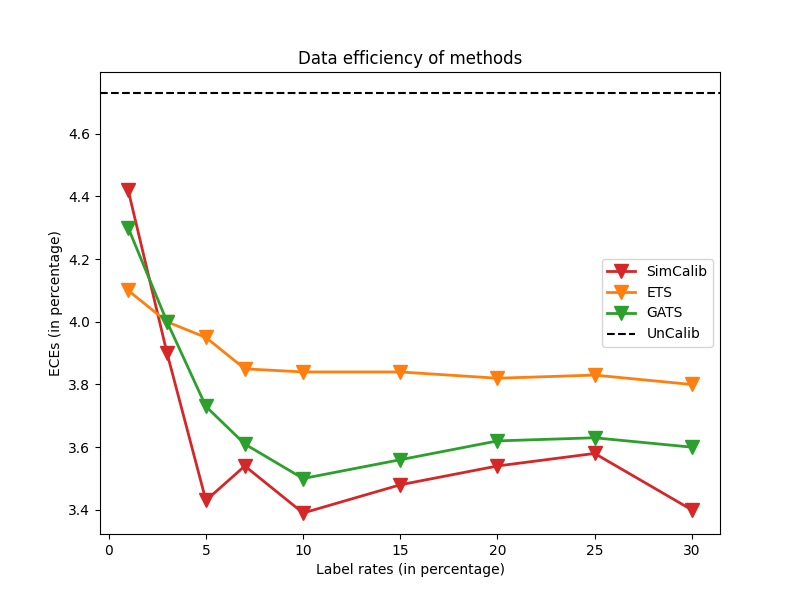}
    \caption{ECEs of different calibrators when applied to GAT.}
    \label{fig:data_eff_gat}
\end{figure}

In Fig.~\ref{fig:data_eff_gat}, we evaluate the calibration performance of SimCalib and two baselines when applied to GAT. The uncalibrated ECE is also plotted as a dashed line as a reference. We omit CaGCN here as its poor calibration results make the performance gaps across other calibrators visually indistinguishable. From the figure, we find a slightly different phenomenon from that shown in Fig.~\ref{fig:efficiency}, in that SimCalib performs worse (although still competitively) than ETS and GATS at extremely low label rate (1\%), but catches up and outperfoms the other baselines immediately. We attribute the inferior performance of SimCalib at low label rate to containing more trainable parameters. SimCalib performs decently even with a small amount of data. Noticeably, although SimCalib takes the form of model ensembling, it significantly outperforms the other ensembling baseline, namely ETS, at most label rates, which verifies the effectiveness of our information-blending design paradigm. The figure validates that our GNN calibrator consistently produces well calibrated confidences for various backbones. Also, SimCalib is expressive for its superior performance at larger label rates.

\subsection{Feature similarity and calibration improvement}
\begin{figure}
    \includegraphics[width=8cm]{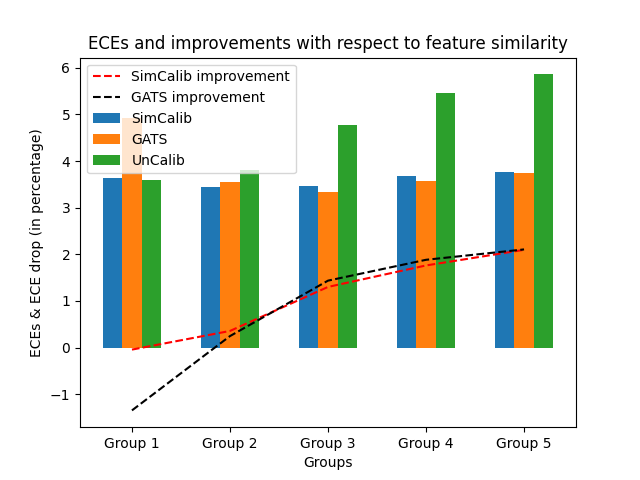}
    \caption{Correlation between feature similarity and calibration improvements for GAT backbone.}
    \label{fig:feature_sim_gat}
\end{figure}

In Fig.~\ref{fig:feature_sim_gat}, we illustrate the correlation between feature similarity and calibration performance \& improvement for SimCalib and GATS, when applied to a pretrained GAT on CoraFull. The uncalibrated ECE worsens with increasing dissimilarity. Also, both GNN calibrators exhibit higher improvements with more dissimilarity. However, whereas SimCalib and GATS perform similarly across 4 out of the 5 groups, SimCalib outperforms GATS at the extremely low feature-similarity scenario, where GATS tangibly reduces calibration performance compared to the uncalibrated backbone. We believe that the design of feature similarity awareness renders SimCalib robust to feature similarities, as shown in the figure.

\subsection{Reliability visualization}
In this section, we provide reliability diagrams and confidence distributions for SimCalib and uncalibrated backbones so that readers can readily assess the improvements of SimCalib. Clearly, we can see that SimCalib consistently calibrates pretrained GNN classifiers.
\begin{figure*}
    \centering
    \includegraphics[width=16cm]{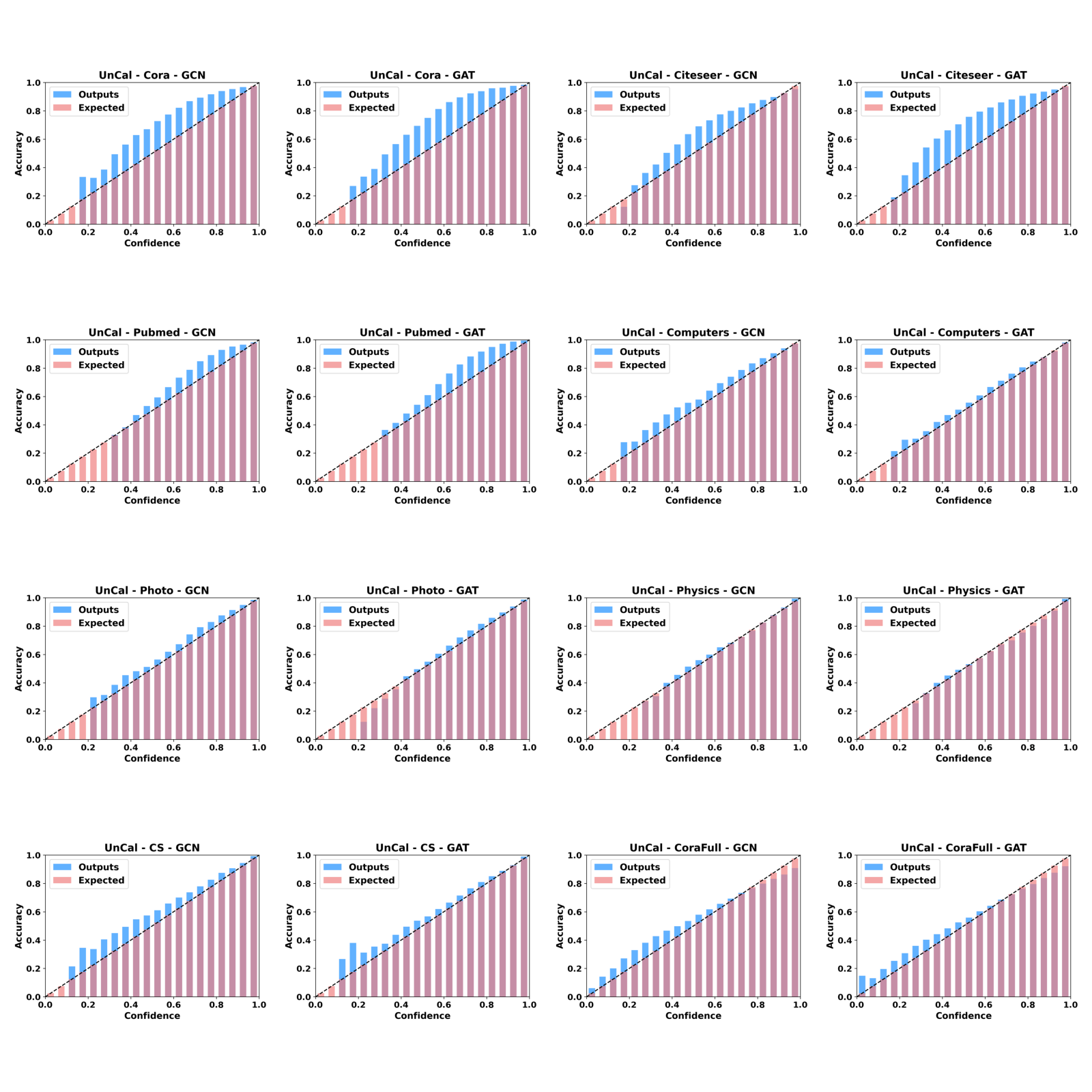}
    \caption{The reliability diagram for uncalibrated GNN classifiers. The horizontal axis represents confidences while the vertical axis is group-wise accuracy. For most datasets, we can see the underconfidence problem of GNNs.}
\end{figure*}
\begin{figure*}
    \centering
    \includegraphics[width=16cm]{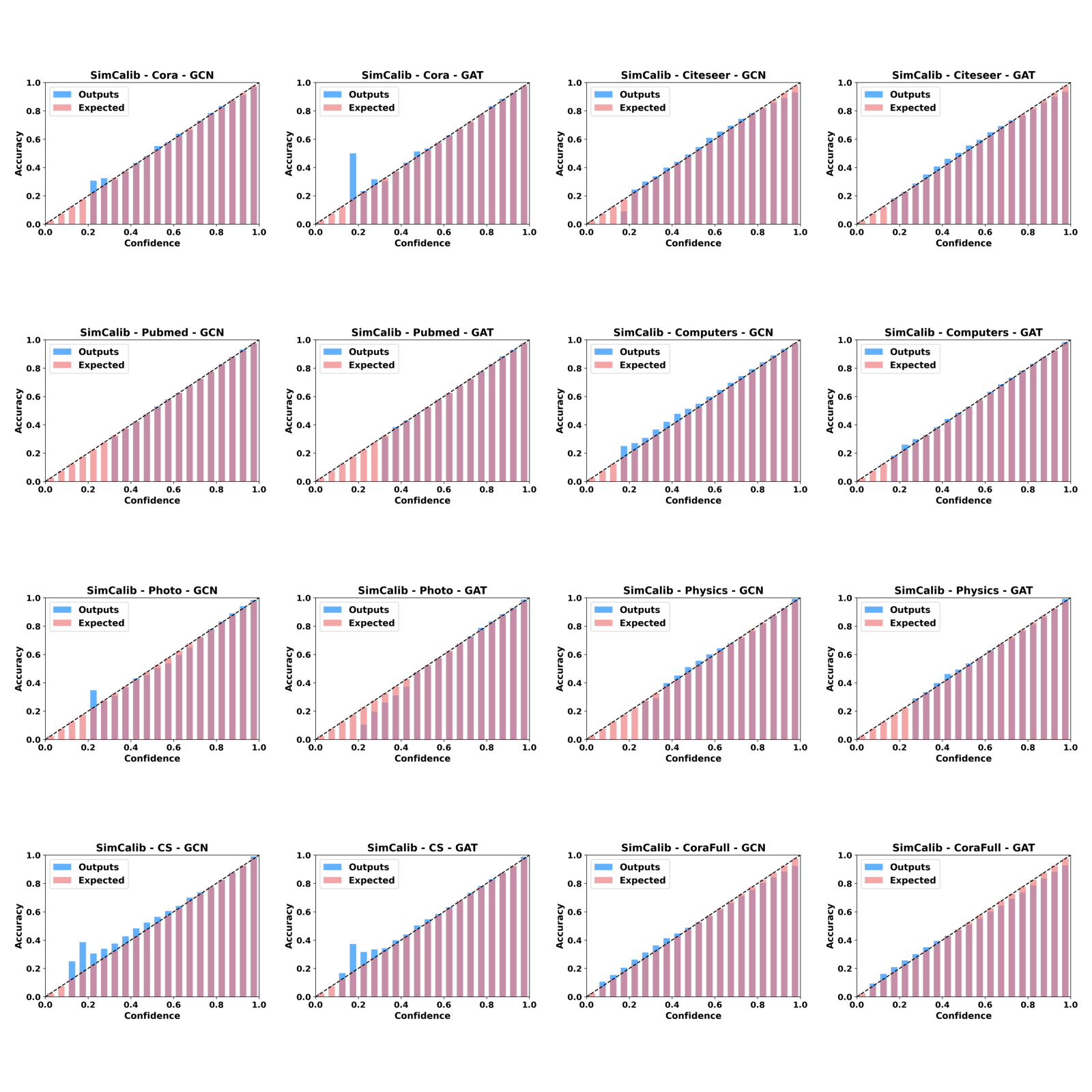}
    \caption{The reliability diagram for uncalibrated GNN classifiers. The horizontal axis represents confidences while the vertical axis is group-wise accuracy. Compared with the previous diagram, one can verify the effectiveness of SimCalib.}
\end{figure*}
\begin{figure*}
    \centering
    \includegraphics[width=16cm]{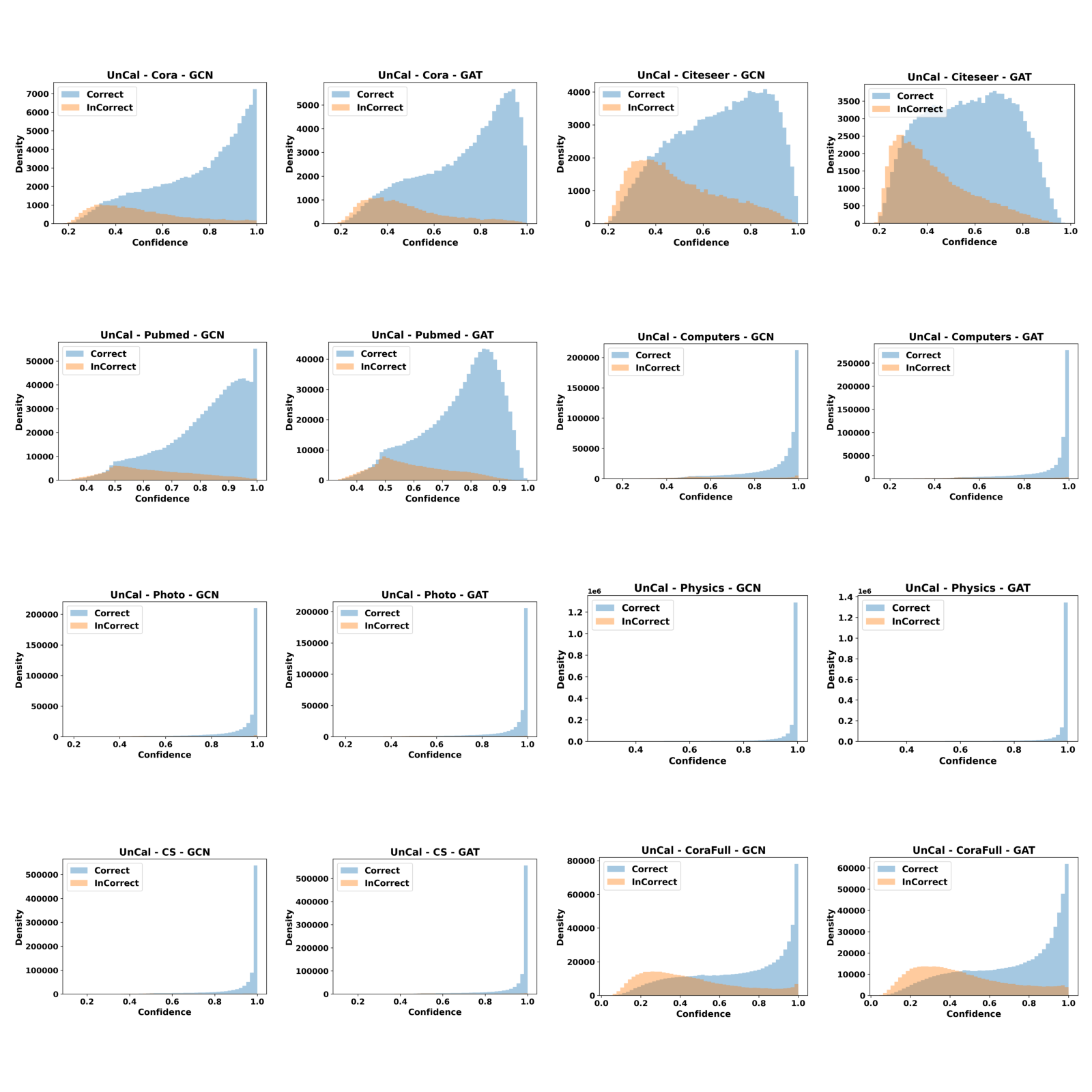}
    \caption{The confidence distributions for uncalibrated GNN classifiers. The horizontal axis represents confidences while the vertical axis is sample density.}
\end{figure*}
\begin{figure*}
    \centering
    \includegraphics[width=16cm]{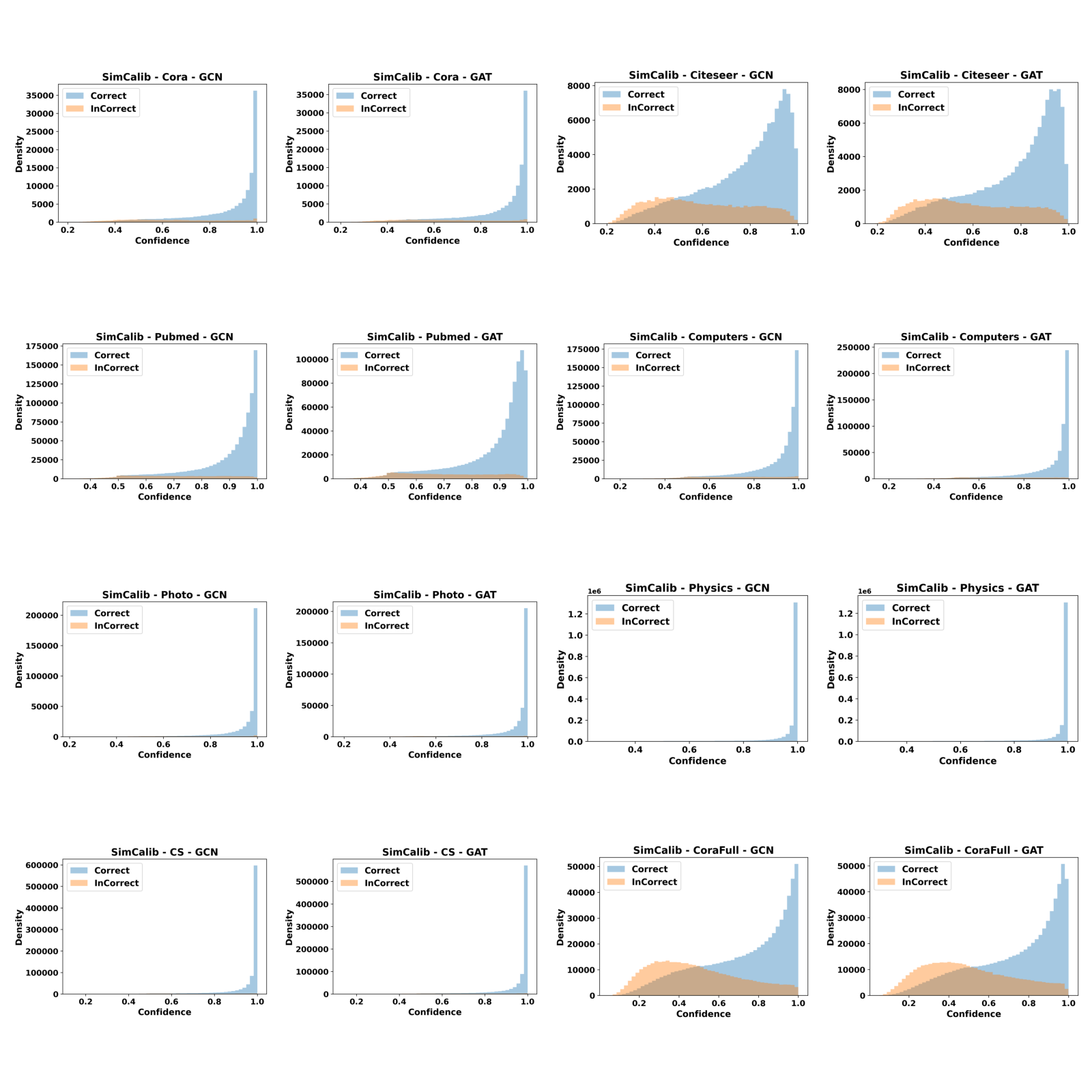}
    \caption{The confidence distributions for SimCalib-calibrated classifiers. The horizontal axis represents confidences while the vertical axis is sample density.}
\end{figure*}

\end{document}